\documentclass[letterpaper]{article} 
\usepackage{aaai23}  
\usepackage{times}  
\usepackage{helvet}  
\usepackage{courier}  
\usepackage[hyphens]{url}  
\usepackage{graphicx} 
\usepackage{natbib}  
\usepackage{caption} 
\usepackage{algorithm}
\usepackage{algorithmic}

\usepackage{xspace}
\usepackage{xcolor}
\usepackage{amsmath,amsfonts}
\usepackage{array}
\usepackage[caption=false,font=normalsize,labelfont=sf,textfont=sf]{subfig}
\usepackage{textcomp}
\usepackage{stfloats}
\usepackage{verbatim}
\usepackage{cite}
\usepackage{microtype}
\usepackage{amssymb}
\usepackage{booktabs} 
\usepackage{multirow}
\usepackage{enumitem}
\usepackage{bbold}
\usepackage{hyperref}

\newcommand{\cmark}{$\surd$}%
\newcommand{\xmark}{$\times$}%
\newcommand{\model}{VLTinT\xspace}
\newcommand{\encoder}{VL Encoder\xspace}
\usepackage{newfloat}
\usepackage{listings}

\usepackage{newfloat}
\usepackage{listings}
\DeclareCaptionStyle{ruled}{labelfont=normalfont,labelsep=colon,strut=off} 
\floatstyle{ruled}
\newfloat{listing}{tb}{lst}{}
\floatname{listing}{Listing}
\title{VLTinT: Visual-Linguistic Transformer-in-Transformer \\ for Coherent Video Paragraph Captioning}
\author{Kashu Yamazaki\equalcontrib\textsuperscript{\rm 1}, 
Khoa Vo\equalcontrib\textsuperscript{\rm 1}, 
Sang Truong\textsuperscript{\rm 1}, 
Bhiksha Raj\textsuperscript{\rm 2, \rm 3}, 
Ngan Le\textsuperscript{\rm 1} 
}

\affiliations{
    \textsuperscript{\rm 1} AICV Lab, University of Arkansas, Fayetteville, Arkansas, USA

    \textsuperscript{\rm 2} Carnegie Mellon University, Pittsburgh, Pennsylvania, USA

    \textsuperscript{\rm 3} Mohammed bin Zayed University of AI
}

\usepackage{bibentry}

\begin{document}

\maketitle

\begin{abstract}
Video paragraph captioning aims to generate a multi-sentence description of an untrimmed video with several temporal event locations in coherent storytelling. 
Following the human perception process, where the scene is effectively understood by decomposing it into visual (e.g. human, animal) and non-visual components (e.g. action, relations) under the mutual influence of vision and language, we first propose a visual-linguistic (VL) feature. In the proposed VL feature, the scene is modeled by three modalities including (i) a global visual environment; (ii) local visual main agents; (iii) linguistic scene elements. We then introduce an autoregressive \emph{Transformer-in-Transformer (TinT)} to simultaneously capture the semantic coherence of intra- and inter-event contents within a video. Finally, we present a new \emph{VL contrastive loss function} to guarantee learnt embedding features are matched with the captions semantics. Comprehensive experiments and extensive ablation studies on ActivityNet Captions and YouCookII datasets show that the proposed Visual-Linguistic Transformer-in-Transform (VLTinT) outperforms prior state-of-the-art methods on accuracy and diversity. Source code is made publicly available at: \url{https://github.com/UARK-AICV/VLTinT}.
\end{abstract}

\section{Introduction}
Video captioning is a task of automatically generating a caption for a video. An important branch of video captioning is dense video captioning (DVC) \cite{krishna2017dense}, which requires to generate a list of temporal event proposals and the associated sentence description of each event to form a coherent paragraph description of a video.
As a simplified version of DVC, video paragraph captioning (VPC) \cite{park2019adversarial} concentrates on generating better paragraph captions given a set of event segments in a video, which eases the requirement of event proposal generation. In general, a VPC model consists of two main components: an encoder to represent each event segment as a feature; and a decoder to generate captions while maintaining the consistency within each event and the coherence among all sentences of the generated paragraph. 


\begin{figure}[!thb]
    \centering
    \includegraphics[width=\linewidth]{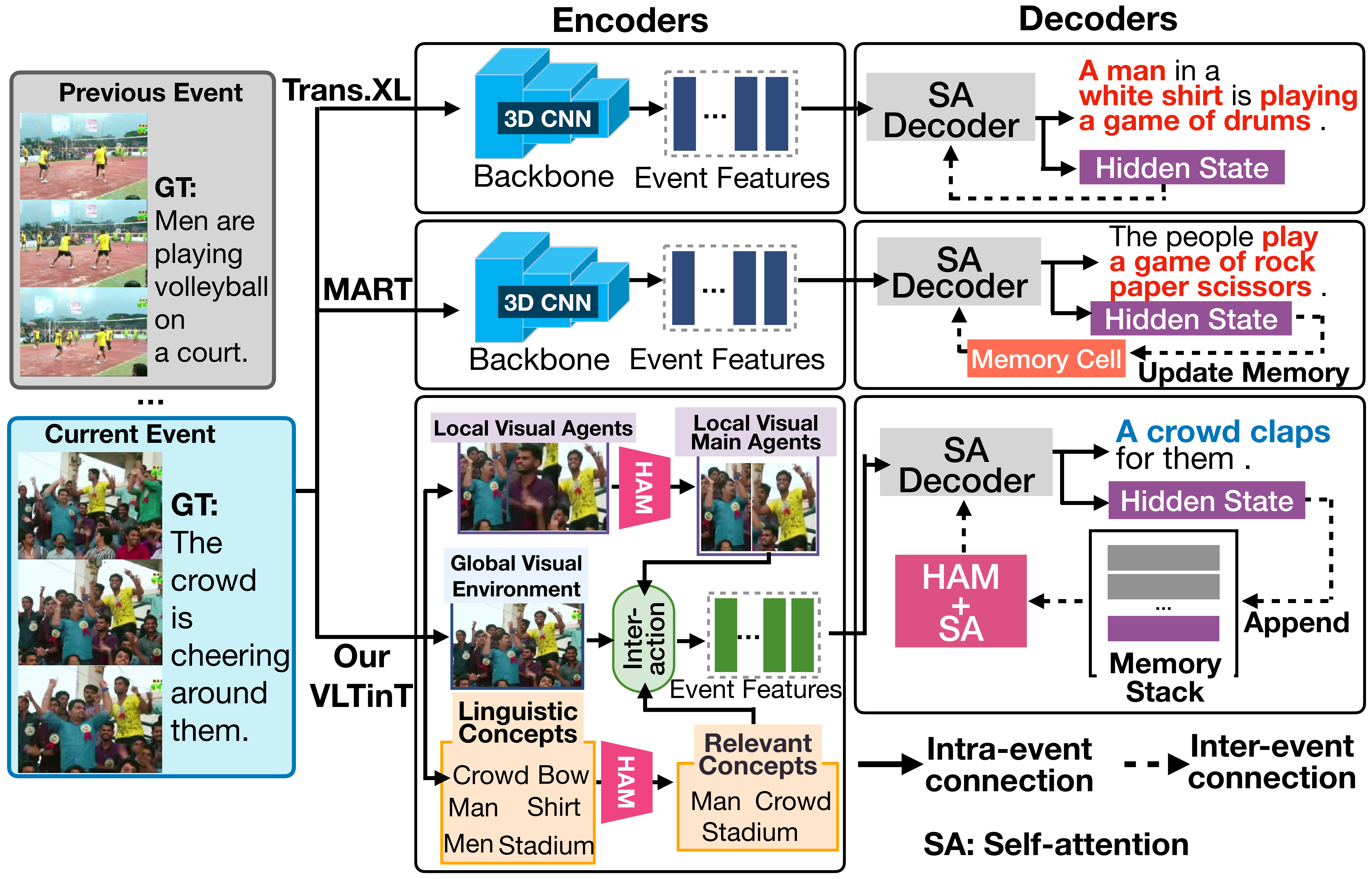}
    \caption{A high-level comparison between our \model and recent SOTA VPC methods. In the \textbf{encoder}, both Transformer-XL \cite{dai2019transformer} and MART \cite{lei2020mart} encode visual features by applying 3D CNN-based backbone network whereas our \model encodes visual-linguistic feature by (i) global visual environment, (ii) local visual main agents, (iii) linguistic scene elements, and a fusion mechanism. In the \textbf{decoder}, Transformer-XL uses recurrence to address context fragmentation, MART uses a highly summarized memory to remember history information whereas we propose to utilize a transformer to model the contextual dependencies at both intra- and inter-levels. In the generated captions, \textbf{\textcolor{red}{red text}} indicates mistaken words, and \textbf{\textcolor{blue}{blue text}} indicates distinct expressions.}
    \label{fig:compare_model}
    \vspace{-2mm}
\end{figure}

Videos contain rich semantic knowledge of multiple modalities, e.g., vision, text, speech, and non-speech audio. Understanding a video involves multiple factors such as a single human actor, group of human actors, non-human actor, and phenomenon. 
Examples of non-human actors and phenomena performing action include dog chasing, car running, and cloud floating. 
The existing VPC approaches \cite{zhou2018end, dai2019transformer, lei2020mart} employ CNN-based networks as a black-box to encode the video feature, which could overlook the contributions of various modalities in the semantic contents of a video.
We observe that human perception involves the \emph{interaction of vision and language} and propose \emph{VL Encoder} to resolve this above challenge. Our \encoder is based on two observations: language influences the basic perceptual process, affecting performance on tasks that might seem to be wholly perceptual in nature \cite{Lupyan2020}; and video content is effectively understood by the combination of agents/actors and the surrounding environment \cite{KhoaVo_ICASSP, KhoaVo_Access, vo2021aei}. 

Our \encoder consists of three modalities: (i) global visual environment representing the overall surrounding scene, (ii) local visual main agents representing the human agents committing events, and (iii) linguistic scene elements captioning descriptive details of both visual and non-visual elements; and a fusion module modeling the interaction of those features and combine them into a unified representation. Besides, to only focus on the main agents who actually contribute to the event as well as the most relevant scene elements of the event, we make use of a Hybrid Attention Mechanism (HAM) following \cite{vo2021aei}.

In VPC, each event is described by one sentence, and they all should logically follow each other. Thus, two kinds of dependencies have to be modeled in VPC, i.e., intra- and inter-event dependencies. In the early days of development, RNN-based models were applied to build the caption generator to model intra-event coherency \cite{xiong2018move, park2019adversarial}. Recently, Transformer-based models have proven to be more effective in generating captions \cite{dai2019transformer, lei2020mart, ging2020coot}. However, in \cite{zhou2018end}, each event is decoded independently and inter-event coherency is not taken into account. This limitation is later addressed as a context fragmentation \cite{dai2019transformer} and RNN-based memory \cite{lei2020mart, ging2020coot}. However, none of the existing work leverages the success of the transformer in modeling  inter-event coherence.
To pose this challenge, we propose a novel \emph{Transformer-in-Transformer architecture (TinT Decoder)}. To the best of our knowledge, TinT Decoder is the first fully transformer network for VPC, which simultaneously models both intra- and inter-event in an end-to-end framework. The network comparison between our VLTinT and the existing SOTA VPC approaches is in Fig. \ref{fig:compare_model}. Furthermore, most prior VPC work makes use of maximum likelihood estimation (MLE) loss to train the model.
However, MLE loss does not guarantee that the learnt event embedding features intimately represent the groundtruth captions. Thus, we introduce a novel \emph{VL contrastive loss}, to maintain the learning of both visual and linguistic semantics during training without adding additional computational costs.
The \encoder along with TinT Decoder comprises a novel method, termed \emph{Visual-Linguistic Transformer-in-Transformer (VLTinT)}. The main contributions of this paper are summarized as follows:

\begin{itemize} [noitemsep,topsep=0pt]
\item A novel \encoder, which represents the video content by separately modeling (i) global visual feature, (ii) local visual main agents, and (iii) linguistic scene elements; and their interactions.
\item A novel TinT Decoder to simultaneously model intra- and inter-event dependencies in an end-to-end fashion producing a coherent paragraph. 
\item A novel VL contrastive loss function to better align both visual and linguistic information. 
\item A comprehensive experiment and an extensive ablation study on the popular VPC benchmarking datasets ActivityNet Captions and YouCookII.
\end{itemize}


\begin{figure*}[!t]
  \centering
  \includegraphics[width=0.85\textwidth]{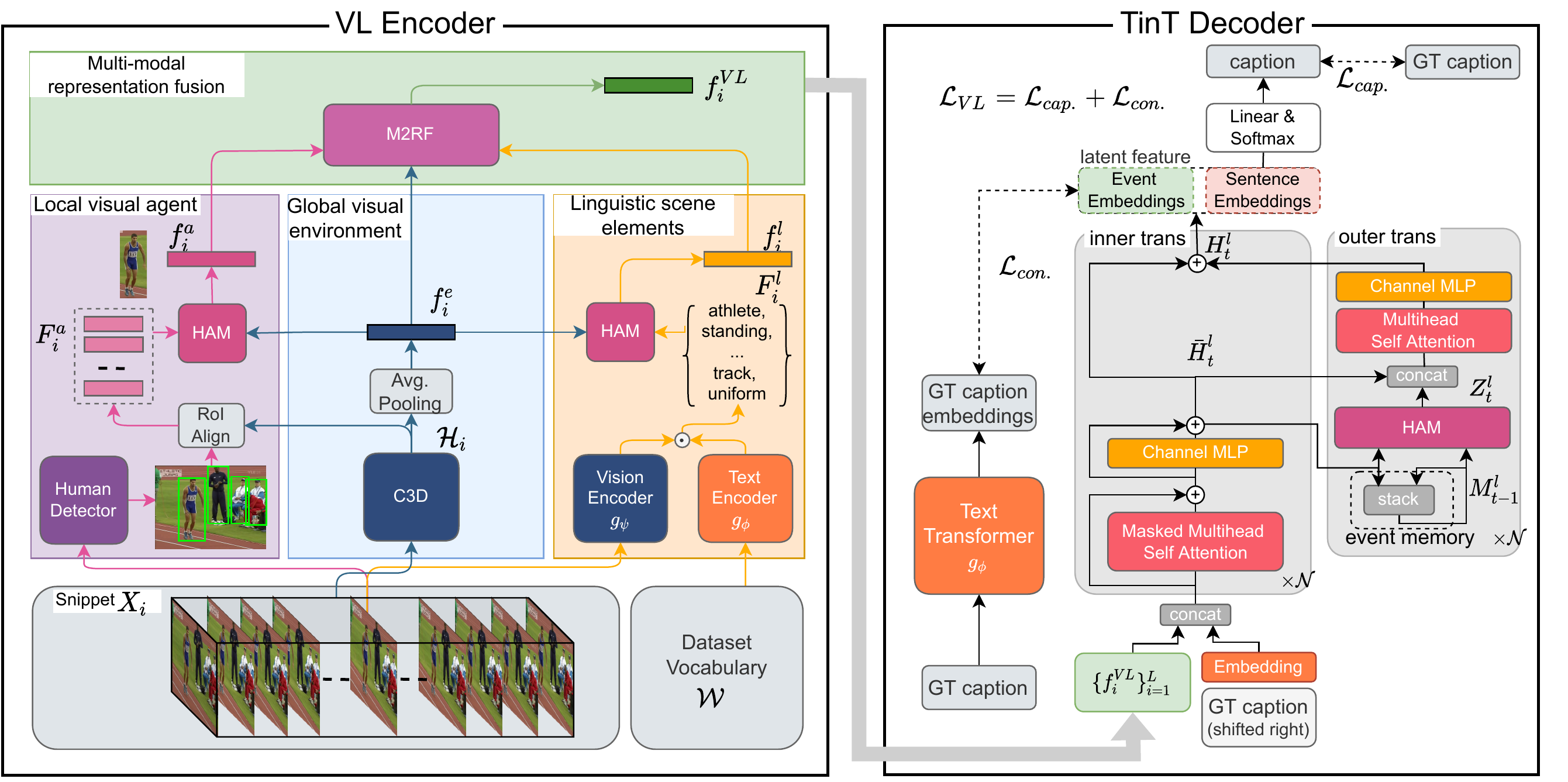}
\caption{Overall network architecture of our proposed \model, which contains two modules, i.e., \encoder and TinT Decoder. (Left) \encoder: given a snippet $X_i$, the \encoder simultaneously extracts local visual features from main agents, global visual features from the environment, and linguistic relevant scene elements; and models interaction between those three modalities through our M2RF module. (Right) TinT Decoder: the canonical transformer encoder is extended by an autoregressive outer transformer that can selectively access the $1^{st}$ to ${t-1}^{th}$ hidden states, which are stored in the event memory, at the $t^{th}$ event captioning step.}
\label{fig:overall}
\vspace{-4mm}
\end{figure*}

\section{Related Works}

\subsection{Dense Video Captioning} 
In general, video captioning can be divided into either single sentence \cite{pasunuru2017multi, wang2019vatex} for short videos or multiple sentences \cite{wang2020event} for long and untrimmed videos. DVC belongs to the second category and it has emerged as a multitask problem that combines event localization and event captioning to generate an informative caption for such videos. DVC can be implemented by visual feature only \cite{krishna2017dense, li2018jointly, zhou2018end, mun2019streamlined, deng2021sketch} or multimodal features such as audio \cite{rahman2019watch}, speech \cite{shi2019dense, iashin2020multi}, and both \cite{iashin2020multi}. Our VPC method shares a common setup with DVC with the multimodal feature. Our feature is encoded using both vision and language modalities to better extract contextual scene representation.

\subsection{Video Paragraph Captioning} 

\cite{zhou2018end} first introduced the transformer to the VPC task known as Vanilla Transformer, where each event is decoded individually without knowing the coherence between sentences. 
To address this limitation, \cite{lei2020mart} modified the Transformer-XL \cite{dai2019transformer} and proposed MART. 
MART decodes the caption to learn word-level dependencies by a transformer while modeling the paragraph coherence based on GRU \cite{chung2014empirical}.  Different from the existing VPC methods which utilize pre-trained backbone networks to extract feature, \cite{yamazaki2022vlcap} inherits the merits of both vision and language models and proposed VLCap. However, all previous works are RNN-based and limited in capturing long-range dependencies as well as suffers from the problem of gradient vanishing \cite{pascanu2013difficulty}. 
In this work, we leverage a transformer to simultaneously model the long-range dependencies between words (i.e., intra-event) and sentences (i.e., inter-event). 

\subsection{Transformer Models}
Transformer \cite{vaswani2017attention} and Vision Transformer (ViT) \cite{dosovitskiy2020image} have recently attracted significant interest in the research community. ViT applies a pure transformer to the visual recognition task by treating the image as a composition of $16\times16$ local patches. 
\cite{Han2021} presents TNT to further divide them into smaller $4\times4$ patches. Specifically, TNT is built with a non-autoregressive inner and outer transformer, which deal with local sub-patches and sequence of local sub-patches, respectively. NesT \cite{Zhang2021} proposes an alternative approach to model local and global information by nesting the canonical transformers hierarchically and connecting them with a proposed aggregation function. 
To model temporal coherency of intra- and inter-event, we propose a novel TinT. In our TinT, the outer transformer is designed as an autoregressive structure to model inter-event coherency whereas the inner transformer handles intra-event coherency.

\section{Proposed VLTinT}

Our \model consists of two main modules corresponding to \encoder and TinT Decoder. The \encoder aims to extract VL representation of each event and the TinT Decoder aims to generate a caption of each event while simultaneously modeling intra- and inter-event coherency. Both modules are trained in an end-to-end fashion by our proposed VL loss. 
The over architecture is shown in Fig. \ref{fig:overall}. 

\subsection{Problem Setup}
\label{sec:setup}
In VPC, we are given an untrimmed video $\mathcal{V}=\{v_i\}_{i=1}^\mathcal{|V|}$, where $|\mathcal{V}|$ is the number of frames, and a list of its important events $\mathcal{E}=\{e_i=(e^b_i, e^e_i)\}_{i=1}^{|\mathcal{E}|}$, where $|\mathcal{E}|$ is the number of events within a video and an event $e_i$ is defined by a pair of beginning and ending timestamps $(e^b_i, e^e_i)$. Our objective is to generate a coherent paragraph that matches the ground truth paragraph $\mathcal{P}=\{\textbf{s}_i\}_{i=1}^{|\mathcal{E}|}$ that describes the whole video $\mathcal{V}$. In this setup, $i^{th}$ sentence $\textbf{s}=\{s_1 \dots s_N\}$ that consists of $N$ words is the description of its corresponding event $e_i$.





\subsection{Visual-Linguistic (VL) Encoder}
\label{sec:vl_encode}
Our \encoder is responsible for comprehensively representing each snippet $X_i$ of an event into a representative feature to compose a sequence of snippet features for the decoder.
Given an event $e=(e^b,e^e)$ and its corresponding video frames $\mathcal{V}_e=\{v_i|e^b\leq i \leq e^e\}$, we follow the standard settings from existing works \cite{zhou2018end, lei2020mart, Song2021} and divide $\mathcal{V}_e$ into a sequence of $\delta$-frame \textit{snippets} $\{X_i\}_{i=1}^L$. Each snippet $X_i$ consists of $\delta$ consecutive frames and $\mathcal{V}_e$ has a total of $L=\bigr\lceil \frac{|\mathcal{V}_e|}{\delta} \bigr\rceil$ snippets. The \encoder module encodes each snippet $X_i$ to a VL representation $f_i^{VL}$ as shown in Fig.\ref{fig:overall} (left). Therefore, video segment $\mathcal{V}_e$ is encoded into VL representation  $\{f_i^{VL}\}_{i=1}^L$. 


The \encoder first models a video with the three modalities, (i) global visual environment (ii) local visual main agents (iii) linguistic relevant scene elements, and then fuses them into one representation based on the interactions between them. Given a snippet $X_i$, it is encoded into these three modalities, corresponding to $f_i^e$, $f_i^a$ and $f_i^l$, respectively. The final feature $f_i^{VL}$ representing the interaction is extracted by fusing $f_i^e$, $f_i^a$ and $f_i^l$ through our Multi-modal Representation Fusion (M2RF) module as follows:

\noindent
\textit{(i) Global Visual Environment:}

This modality provides the visual semantic information from the entire spatial scene of input snippet $X_i$. To obtain such target, we adopt a backbone 3D-CNN network \cite{C3D} to $X_i$ to extract feature map $\mathcal{H}_i$ at the last convolutional block of the network. Then, we obtain the global environment visual feature $f^e_i \in\mathbb{R}^{d_\text{emb}}$ by processing $\mathcal{H}_i$ with an average pooling operation to reduce the entire spatial dimension followed by channel MLP. The procedure is summarized as follows:
\begin{equation}
    f^e_i = \text{MLP}_{\theta_e}(\text{Avg.Pooling}(\mathcal{H}_i))
\end{equation}

\noindent
\textit{(ii) Local Visual Main Agents:}

This modality provides the visual features of the main human agents, who actually contribute to the formation of the event being described. Even though most of the events are associated with agents, not all agents committing movements are related to the main content of the event segment.
Using a similar assumption as in \cite{KhoaVo_ICASSP, KhoaVo_Access}, we apply a human detector to the center frame of $X_i$ to obtain the bounding boxes of all human agents. 
Afterward, we align each of the detected bounding boxes $\mathcal{B}_i$ onto the feature map $\mathcal{H}_i$, which is obtained by the previous modality, using RoIAlign \cite{MaskRCNN_ICCV17}. Then, features overlapped by each agent bounding box are averagely pooled into a single feature vector to represent visual information of the agent inside that box. Finally, we obtain a set of local agent visual features $F^a_i\in\mathbb{R}^{N_a\times d_a}$, where $N_a$ and $d_a$ are the number of detected agents and agent embedding dimension, respectively.
Finally, we apply HAM (detailed in Section \ref{sec:aam}) to adaptively select an arbitrary number of main agents from $N_a$ detected agents and extract their mutual relationships to form a unified agent-aware visual feature $f^{a}_i \in \mathbb{R}^{d_\text{emb}}$ as follows:
\begin{align}
 f_i^a = \text{HAM}(\text{MLP}_{\theta_a}(F_i^a), f_i^e)
\end{align}

\noindent
\textit{(iii) Linguistic Relevant Scene Elements:}
\label{sec:ling}

\begin{figure}[t]
\centering
  \includegraphics[width=0.9\linewidth]{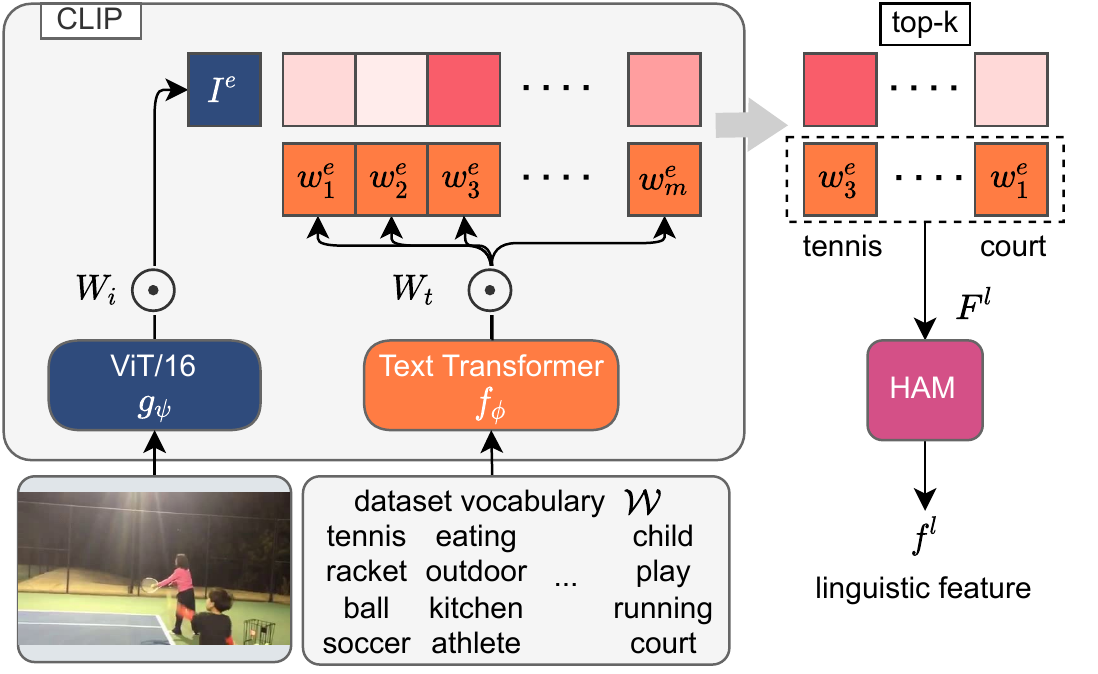}
  \caption{Illustration of relevant scene elements extraction process where ViT/16 and Text Transformer are the pre-trained models from CLIP \cite{radford2021learning}.}
  \label{fig:lin_feat_v2}
 \vspace{-4mm}
\end{figure}

This modality provides additional contextual details of the scene. 
While the two former modalities capture visual information of spatial appearances and temporal motions, their features may overlook some of the scene components because of the spacial reduction in the pooling operation from the feature map $\mathcal{H}_i$. Furthermore, non-visual features could hardly be captured by a normal vision backbone model. 
Recent studies \cite{Patashnik2021styleclip, Yang2021} have shown the extreme zero-shot capability of Contrastive Language-Image Pre-training (CLIP) where the model can estimate the semantic similarity between a set of words and an image. 
Trained on 400 million text-image pairs collected from a variety of publicly available sources on the Internet, CLIP can correlate not only the visual words but also the non-visual words to the given image.
We thus leverage CLIP as a linguistic feature extractor to obtain top $k$ scene elements (i.e., $k$ texts) that are highly correlated with the middle frame of the input snippet $X_i$. 
Specifically, we construct a vocabulary $\mathcal{W} = \{w_1, \dots w_m\}$ based on the groundtruth captions of our training dataset. Each vocabulary $w_i\in \mathcal{W}$ is encoded by a transformer network $f_\phi$ into a text feature $f_i^w$. Let $W_t$ be a text projection matrix pre-trained by CLIP, the embedding text vocabulary is computed as 
\begin{equation}
    w^e = W_t \cdot f_\phi(\mathcal{W}) = W_t \cdot f^w
    \text{where }
    f^w = \{f_i^w\}_{i=1}^{m}.
\end{equation} 
 
Let $W_i$ be an image projection matrix pre-trained by CLIP, the center frame $I$ of the input snippet $X_i$ is first encoded by a pre-trained ViT $g_\psi$ to extract visual feature $f^I$, and then embedded by $W_i$ as below:
\begin{equation}
    I^e = W_i\cdot g_\psi(I) = W_i\cdot f^I
\end{equation}
The pairwise cosine similarities between embedded $I^e$ and $w^e$ are then computed. Top $k$ similarity scores are chosen as linguistic categorical concept features $F_i^l\in \mathbb{R}^{k\times d_l}$. This feature is also subjected to the HAM module to select only the most relevant representative linguistic features and merge them into a single representation $f_i^{l} \in\mathbb{R}^{d_\text{emb}}$ as follows:
\begin{equation}
    f_i^{l} = \text{HAM}(\text{MLP}_{\theta_l}(F_i^l), f_i^e)
\end{equation}
The flowchart of extracting $f_i^{l}$ is illustrated in Fig.\ref{fig:lin_feat_v2}.

\noindent
\textit{(iv) Multi-modal Representation Fusion (M2RF):}

This component aims to fuse features from the three modalities. While concatenation or summation are the two common fusion mechanisms, they treat all modalities equally. To better model the impact of each individual modality, we propose M2RF as a function $g_\gamma$, which takes the features $f_i^e$, $f_i^a$, and $f_i^l$ as its input.
We extract the inter-feature relationships by utilizing a self-attention (Att.) layer \cite{vaswani2017attention} followed by a mean operation. The final representation $f_i^{VL} \in \mathbb{R}^{d_\text{emb}}$ of a given snippet $X_i$ is defined as follows:
\begin{subequations}
\begin{align}
    f_i^{VL} = g_\gamma([f^e_i;f^a_i;f^l_i ]) =\text{mean}(\text{Att.}([f^e_i;f^a_i;f^l_i ])) \label{g_gamma}
\end{align}
\label{eq:m2rf}
\end{subequations}
where $[;]$ represents the concatenation of features in a new dimension, where self-attention is applied on the new dimension and reduced by the mean operation to account for permutation invariance.

\subsection{Transformer-in-Transformer (TinT) Decoder}
\label{sec:TinT}

Inspired by the recent transformer-based vision-language models \cite{Chen2019uniter, lei2020mart}, we adopt the unified encoder-decoder transformer structure as a foundation for the caption generator, i.e., an inner transformer. The inner transformer's input is described as following.
In this setup, video features $\mathcal{F}^{VL}$ is formed by concatenating all $f_i^{VL}$ obtained by applying \encoder into each snippet $X_i$, i.e., $\mathcal{F}^{VL} = \{f_i^{VL}\}^{L}_{i=1}\in \mathbb{R}^{L\times d_{\text{emb}}}$. Textual tokens $\mathcal{F}^{text}$ is encoded by a pre-trained text transformer $g_\phi$ from CLIP and a MLP layer, i.e., $\mathcal{F}^{text} = MLP(g_\phi(\text{Shifted GT text})) \in \mathbb{R}^{N\times d_{\text{emb}}}$, where $N$ is the sequence length of the text tokens. Following \cite{lei2020mart}, learnable token type embeddings $\mathcal{F}^{type} \in \mathbb{R}^{(L+N) \times d_{\text{emb}}}$ are introduced to inform the location of the video and the caption representations. $\mathcal{F}^{type}$ is initialized as 0/1 vectors, i.e., video as 0 and text as 1. For the $t^{th}$ event, an intermediate hidden states $\bar{H}_t^l \in \mathbb{R}^{(L+N) \times d_{\text{emb}}}$ is computed in Eq. \ref{eq:inter1} as canonical inner transformer encoder, where $\Tilde{H}_t^l$ is the internal states after Masked Multihead Self Attention (MSA).

\begin{subequations}
\begin{align}
    H_t^0 &= [\mathcal{F}^{VL}; \mathcal{F}^{text}] +  \mathcal{F}^{type} \in \mathbb{R}^{(L+N) \times d_{\text{emb}}}\\
    \bar{H}_t^l &= \text{MLP}(\Tilde{H}_t^l) + \Tilde{H}_t^l \text{, } \Tilde{H}_t^l= \text{Masked\_MSA}(H_t^l)+H_t^l \label{eq:inter1}
\end{align}
\end{subequations}


While the inner transformer can effectively model intra-event coherency, it cannot handle the contextual relationship of inter-event. To address this limitation, we introduce an autoregressive outer transformer. The outer transformer selectively utilizes the activations of the inner transformer from the previous time steps for generating a coherent paragraph.  
Specifically, we take advantage of HAM to select only the most relevant hidden states of all previous events stored in event memory with respect to the current one.
The outer transformer process is formulated below:
\vspace{-1mm}
\begin{subequations}
\begin{align}
    &M_t^l = [M_{t-1}^l; \bar{H}_t^l ]\label{eq:mem}\\
    &Z_t^l = \text{HAM}(M_{t-1}^l, \bar{H}_t^l)\label{eq:ham_update}\\
    &H_t^l = \text{MLP}(g_\gamma([\bar{H}_t^l; Z_t^l])) + \bar{H}_t^l
    \label{eq:tint_3}
\end{align}
\end{subequations}
 
For the $t^{th}$ event, an intermediate hidden states 
$\bar{H}_t^l$ is stacked to the event memory $M_{t}^l \in \mathbb{R}^{t \times(L+N) \times d_{\text{emb}}}$, where $M_0^l = \varnothing$ as in Eq.~\ref{eq:mem}.  Eq.~\ref{eq:ham_update} computes the context $Z_t^l$ from the previous states of the event memory and the current intermediate hidden states $\bar{H}_t^l$ using HAM. Finally, in Eq.~\ref{eq:tint_3}, the context is integrated with the intermediate hidden states $\bar{H}_t^l$ using $g_\gamma$, which was introduced in Eq.~\ref{g_gamma}, and the hidden states are updated via residual connection. 
After the last layer, video token positions in $H_t^{\mathcal{N}}$ are ignored, and only the text token positions are fed to a feed-forward layer followed by softmax to predict a caption for the $t^{th}$ event. 



\subsection{Hybrid Attention Mechanism (HAM)}
\label{sec:aam}

\begin{figure}[!t]
\centering
  \includegraphics[width=0.9\linewidth]{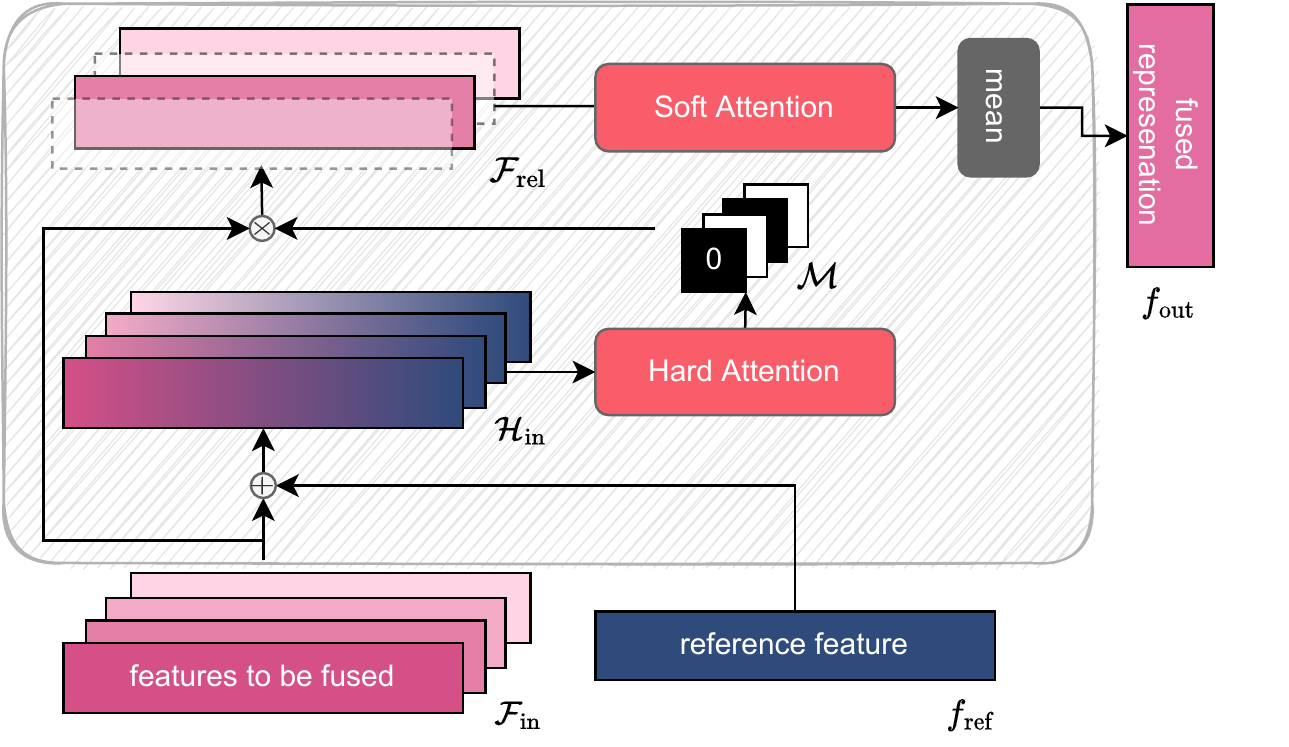}
  \caption{Illustration of HAM. HAM is capable of selecting and representing an arbitrary number of representative features from the input features $\mathcal{F}_{in}$ with a guidance from reference feature $f_{\text{ref}}$. }
  \label{fig:AAM}
\end{figure}

HAM inherits the merits of both hard attention \cite{patro2018differential} and the self-attention \cite{vaswani2017attention} to select a rational number of representative features out of a set of input features and to extract mutual relationships among the sub-set of selected features, respectively, and fuse them into a unified representation.
Denote $\mathcal{F}_{\text{in}} \in \mathbb{R}^{N_{\text{in}} \times d_{\text{in}}}$ and $f_{\text{ref}} \in \mathbb{R}^{d_{\text{in}}}$ as a set of input features and a reference feature, respectively, where $N_{\text{in}}$ is the total number of input features and $d_{\text{in}}$ is the embedding dimension of input and reference features. HAM takes $\mathcal{F}_{\text{in}}$ and $f_{\text{ref}}$ as inputs and compute the most relevant feature $f_{\text{out}}$ as its output. Fig.~\ref{fig:AAM} visualizes the workflow of HAM, which is formulated as follows: 

\vspace{-1mm}
\begin{subequations}
\begin{align}
&\mathcal{H}_{\text{in}}=\mathcal{F}_{\text{in}} \oplus f_{\text{ref}}\label{eq13}\\
&\mathcal{C}=\text{softmax}(||\mathcal{H}_{\text{in}}||_2)\label{eq14}\\
&\mathcal{M} = \mathcal{C} > \frac{1}{N_{in}}\label{eq15}\\
&f_{\text{out}} = g_\gamma(\mathcal{F}_{\text{in}} \odot \mathcal{M})\label{eq17}
\end{align}
\end{subequations}
where Eq.~\ref{eq13} broadcasts the reference feature $f_{\text{ref}}$ into every input feature of $\mathcal{F}_{\text{in}}$ by element-wise addition to form a set of hidden features $\mathcal{H}_{\text{in}}$. Then Eq.~\ref{eq14} computes $L2$-norm value of each hidden feature in $\mathcal{H}_{\text{in}}$, and re-scales all values to be summed up to $1.0$ using softmax, this makes a corresponding value of each input feature represent its likelihood to be a representative feature of the entire input features set. Afterward, Eq.~\ref{eq15} defines an adaptive threshold that 
creates a mask $\mathcal{M}$ to 
filter out features that are predicted to be non-representative. Finally, Eq.~\ref{eq17} uses self-attention to extract mutual relationships among selected input features and average them into a single representation feature $f_{\text{out}} \in \mathbb{R}^{d_{\text{in}}}$ of the entire input features set.

\subsection{Visual-Linguistic (VL) contrastive Loss}
\label{sec:VL_loss}
Typically, the existing VPC methods exploit the MLE loss to train their models. The MLE loss serves the objective of increasing the likelihood of predicted captions to be matched with the groundtruths. However, it is unable to address the question of how well the learnt event embedding features represent the groundtruth captions. To this end, we leverage the recent advantages of contrastive learning \cite{wu2018unsupervised, chen2020simple} and propose $\mathcal{L}_{con}$ to pull all snippets of the same event and push snippets of different events. Our VL Loss consists of two terms corresponding to captioning loss ($\mathcal{L}_{cap.}$) and a contrastive contextual loss ($\mathcal{L}_{con.}$). While $\mathcal{L}_{cap.}$ aims to decode captions that match with groundtruths, $\mathcal{L}_{con.}$ guarantees the learnt latent features are close to the semantic information encoded in the groundtruth captions.

\noindent\textbf{Captioning Loss $\mathcal{L}_{cap.}$}:  
Kullback–Leibler (KL) divergence is commonly utilized
to minimize the divergence between empirical distribution $p(\mathbf{s}|\mathcal{V}_e)$ and 
predicted distribution $p_\theta(\mathbf{s}|\mathcal{V}_e)$ for a video segment $\mathcal{V}_e$. 
However, this objective easily makes the captioning model overfit high-frequency tokens and phrases, which results in repetitive phrases. In order to enhance the smoothness of the predicted sentence, a regularization term $\tau$ is introduced to the training objective with hyper-parameter $\lambda$ as:

\vspace{-2mm}
\begin{equation}
\begin{split}
     \theta^* 
     &= \mathop{\mathrm{argmin}}_\theta \mathbb{E}_{\mathbf{s}\sim  p(\mathbf{s})}\left[\log{\left(\frac{p(\mathbf{s})}{p_\theta(\mathbf{s})}\right)} + \lambda\tau(\mathbf{s})\right]
\end{split}
\end{equation}
The second term $\tau$ imposes a token-level high-frequency penalties as \cite{Song2021}. Based on the observation that the model tends to generate words that have been generated before, we penalize the previously appeared words in the regularization term:  
\begin{equation}
    \tau(\mathbf{s}) = -\frac{1}{N}\sum_{i=1}^N\sum_{c\in \{s | s_{<i}\}}\log{(1-p_\theta(c|s_{<i}, \mathcal{E}))}
\end{equation}

where $c$ is the candidate word at $n$ to be penalized.

\noindent Our $\mathcal{L}_{cap.}$ is defined as follows:
\begin{equation}
    \mathcal{L}_{cap.} = -\frac{1}{N}\sum_{i=1}^{N}(\log{p_\theta(s_i | s_{<i}, \mathcal{V}_e)}) + \lambda\tau(\mathbf{s})
\end{equation}

where $\theta$ is the model parameters, $s_{1:N}$ is the target ground truth sequence. 


\begin{table*}[!t]
\centering
\caption{Performance comparison of \model with other SOTA models on ActivityNet Captions \textit{ae-val}. $\dag$ denotes results by us. }
\resizebox{\linewidth}{!}{
\begin{tabular}{l|l|l|cccc|cc}
\toprule
\label{tab:anet_val}
Methods & Venue &Input & B\@4 $\uparrow$ & M $\uparrow$&  C $\uparrow$ & R $\uparrow$  & Div\@2 $\uparrow$ & R\@4 $\downarrow$ \\ \hline

Vanilla Trans. \cite{zhou2018end} & CVPR & Res200/Flow & 9.75 & 15.64& 22.16 & 28.90$^\dag$ & \underline{77.40}$^\dag$ & 7.79 \\ 


AdvInf \cite{park2019adversarial} & CVPR & C3D/Object & 10.04 & \underline{16.60} & 20.97 & -- & -- &  5.76 \\ 
GVD \cite{zhou2019grounded} & CVPR  & Res200/Flow/Object &  11.04 & 15.71 &  21.95 & -- & -- & 8.76\\ 

Trans.-XL \cite{dai2019transformer} & ACL & Res200/Flow & 10.39 & 15.09 & 21.67 & 30.18$^\dag$ & 75.96$^\dag$ & 8.54 \\ 
Trans.-XLRG \cite{lei2020mart} & ACL & Res200/Flow & 10.17 & 14.77 & 20.40 & -- & -- & 8.85 \\
MART \cite{lei2020mart} & ACL & Res200/Flow &  10.33 & 15.68 & 23.42 &  \underline{30.32}$^\dag$ & 75.71$^\dag$ & \underline{5.18} \\
PDVC \cite{wang2021end} & ICCV & C3D/Flow & \underline{11.80} &  15.93 &  \underline{27.27}  & -- & -- & --\\

\hline
\textbf{\model} (ours) & -- & C3D/Ling & \textbf{14.93} & \textbf{18.16} & \textbf{33.07} & \textbf{36.86} & \textbf{77.72} & \textbf{4.87} \\
\bottomrule
\end{tabular}}
\end{table*}
\begin{table*}[!ht]
\centering
\caption{Performance comparison of \model with other SOTA models on ActivityNet Captions \textit{ae-test}. $\dag$ denotes results by us.} 
\resizebox{\linewidth}{!}{
\begin{tabular}{l|l|l|cccc|cc}
\toprule
\label{tab:ex2}
Methods & Venue & Input & B\@4 $\uparrow$ & M $\uparrow$&  C $\uparrow$ & R $\uparrow$ & Div\@2 $\uparrow$  & R\@4 $\downarrow$\\ \hline
Vanilla Trans. \cite{zhou2018end} & CVPR & Res200/Flow & 9.31 & 15.54&  21.33& 28.98$^\dag$ &  77.29$^\dag$ & 7.45\\
Trans.-XL \cite{dai2019transformer}  & ACL & Res200/Flow & 10.25 & 14.91 & 21.71 & 30.25$^\dag$ & 76.17$^\dag$ &8.79 \\ 
Trans.-XLRG \cite{lei2020mart}  & ACL & Res200/Flow & 10.07 & 14.58 & 20.34 & -- & --&  9.37 \\
MART \cite{lei2020mart} & ACL   & Res200/Flow &  9.78 & 15.57 & 22.16 & \underline{30.85}$^\dag$ & 75.69$^\dag$ & 5.44 \\
MART$^{\text{COOT}}$ \cite{ging2020coot}  & NIPS  & COOT & 10.85 &  \underline{15.99} & \underline{28.19} & -- & -- &  6.64 \\
Memory Trans. \cite{Song2021}  & CVPR  & I3D &  \underline{11.74} & 15.64 & 26.55 & -- & \textbf{83.95} & \textbf{2.75} \\
\hline
\textbf{\model} (ours) & -- & C3D/Ling & \textbf{14.50} & \textbf{17.97} & \textbf{31.13} & \textbf{36.56} & \underline{77.72} &  \underline{4.75}  \\
\bottomrule
\end{tabular}}
\label{tab:anet_test}
\end{table*}

\noindent\textbf{Contrastive Contextual Loss $\mathcal{L}_{con.}$}:
We propose $\mathcal{L}_{con.}$ to optimize the latent feature of the input event to be highly correlated with its groundtruth description. This loss function implicitly encourages our \model to learn better representations of the events and enhance its overall performance without extra computational cost. 

Specifically, $\mathcal{L}_{con.}$ processes the entire mini-batch of training examples $\mathcal{B}=\{(\mathcal{V}_b, \textbf{s}_b)\}_{b=1}^{|\mathcal{B}|}$, where $\mathcal{V}_b$ is a set of snippets within the same event and $\textbf{s}_b$ is its corresponding groundtruth description sentence. 
On the one hand, video snippets in $\mathcal{V}_b$ are processed through our proposed \model to obtain the event embeddings, which corresponds to the video token position $\mathcal{F}^{\mathcal{N}}_b\in \mathbb{R}^{L\times d_\text{emb}}$ of the final hidden state $H_b^{\mathcal{N}}$. On the other hand, we process each groundtruth caption sentence $\textbf{s}_b$ through the transformer $g_\phi$ of CLIP \cite{radford2021learning} to obtain a representation feature $f^\mathcal{T}_b \in \mathbb{R}^{d_\text{emb}}$. Then, $\mathcal{L}_{con.}$ processes $\mathcal{F}^{\mathcal{N}}_b$ and $f^\mathcal{T}_b$ as follows:

\begin{subequations}
\vspace{-3pt}
\begin{align}
    \mathcal{L}_{con.} =& -\sum_{b_1=1}^{|\mathcal{B}|} \sum_{b_2=1}^{|\mathcal{B}|} [\mathbb{1}_{b_1=b_2}\log (e^\rho( f_{b_1}^\mathcal{N} \cdot f_{b_2}^\mathcal{T})) \nonumber\\
    & + (1-\mathbb{1}_{b_1=b_2})(1-\log (e^\rho( f_{b_1}^\mathcal{N} \cdot f_{b_2}^\mathcal{T}))) ]
\end{align}
\end{subequations}

where $f_b^\mathcal{N} = \text{mean}(\mathcal{F}^{\mathcal{N}}_b)$. $\mathbb{1}_{b_1=b_2}$ returns 1 when samples come from the same event, i.e., $b_1=b_2$ and 0 when samples come from the different events i.e., $b_1\neq b_2$. $\rho$ is a learnable temperature parameter initialized as $\log(1/0.07)$, to prevent scaling of the dot product values and stabilize the training. 


\noindent Finally, our proposed VL contrastive loss $\mathcal{L}_{VL}$ is defined as:
\begin{equation}
    \mathcal{L}_{VL} = \mathcal{L}_{cap.} + 
    \mathcal{L}_{con.}
\end{equation}

\begin{table*}[!hbt]
\centering
\caption{Performance comparison of \model with other SOTA models on YouCookII validation set.}
\begin{tabular}{l|l|l|cccc|c}
\toprule
Methods & Venue & Input & B@4 $\uparrow$ & M $\uparrow$&  C $\uparrow$ & R $\uparrow$ & R@4 $\downarrow$\\ \hline
Vanilla Trans.\cite{zhou2018end} & CVPR  & Res200/Flow & 4.38 & 11.55 & 38.00 & -- & --\\
MART \cite{lei2020mart} & ACL  &  Res200/Flow & 8.00   & 15.90 & 35.74 & -- & \underline{4.39} \\
MART$^{\text{COOT}}$ \cite{ging2020coot}  & NIPS & COOT & \textbf{9.44} & \textbf{18.17} & \underline{46.06} & -- & 6.30\\
\hline
\textbf{\model} (ours) & -- & C3D/Ling & \underline{9.40} & \underline{17.94} & \textbf{48.70} & \textbf{34.55} & \textbf{4.29}  \\
\bottomrule
\end{tabular}
\label{tab:youcook}
\end{table*}

\section{Experiments}

\subsection{Datasets and Metrics}

We benchmark \model on two popular datasets ActivityNet Captions~\cite{krishna2017dense} and YouCookII~\cite{zhou2018towards}. ActivityNet Captions consists of 10,009 training and 4,917 validation videos. 
On average, there are 3.65 event segments for each video. 
We follow the previous works \cite{lei2020mart} to split the original validation set into two subsets: \textit{ae-val} with 2,460 videos for validation and \textit{ae-test} with 2,457 videos for test. YouCookII contains 1,333 training and 457 validation videos. 
On average, there are 7.7 event segments for each video. We report our results on the validation sets. 


We evaluate the performance on four standard metrics, i.e., BLEU@4 (B@4) \cite{papineni2002bleu}, METEOR (M) \cite{denkowski2014meteor}, CIDEr (C) \cite{vedantam2015cider}, ROUGE (R) \cite{lin2004rouge}. Whereas to benchmark the diversity of generated captions, we use two diversity metrics, including 2-gram diversity (Div@2) \cite{div} and 4-gram repetition (R@4) \cite{xiong2018move}.

\subsection{Implementation Details}
To extract visual features of the environment, we use C3D \cite{C3D} pre-trained on Kinetics-400 \cite{Kinetics} as the backbone network. 
The agent visual feature is extracted by Faster-RCNN \cite{FasterRCNN} that is pre-trained on the COCO dataset \cite{cocodataset}. To extract the linguistic scene element features, we employ CLIP \cite{radford2021learning} ViT-B/16 model made publically available by OpenAI.
We set the hidden size to 768, the number
of transformer layers to 3, and the number of attention heads to 12.
Adam optimizer was used to train \model with an initial learning rate of 1e-4, $\beta_1=0.9$, $\beta_2=0.999$, $L_2$ weight decay of 0.01, and learning rate warmup over the first 5 epochs. During the training, we use the label smoothing with a value of 0.1 and $\lambda=0.1$. We ran the experiment on a single NVIDIA RTX 3090 (24GB) GPU. 

\subsection{Qualitative Analysis}
\begin{figure}[!hbt]
\centering
  \includegraphics[width=\linewidth]{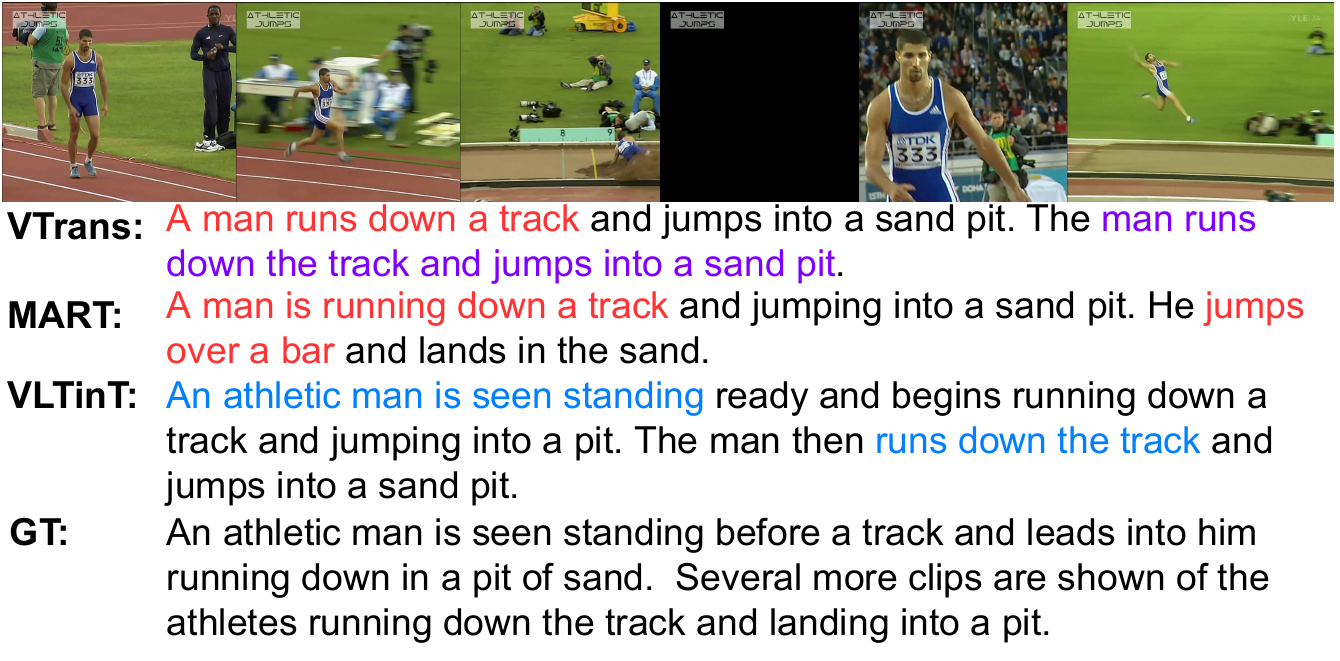}
  \caption{Qualitative comparison on ActivityNet Captions \textit{ae-test} split. \textcolor{red}{Red text} indicates the captioning mistakes, \textcolor{violet}{purple text} indicates repetitive patterns, and \textcolor{blue}{blue text} indicates some distinct expressions.}
  \label{fig:qualitative}
\end{figure}

Fig.\ref{fig:qualitative} shows comparison between \model and Vanilla Transformer (VTrans) \cite{zhou2018end} and MART \cite{lei2020mart}. Overall, \model can generate more descriptive captions with fine-grained details. In particular, we noticed that VTrans and MART are prone to use high-frequency words for their caption, while \model can use expressive but less frequently appearing words, e.g., 
"A man" vs. "An athletic man" in the example. We attribute this improvement to our \encoder, which incorporates relative scene elements. We further observe a caption repetitiveness problem in VTrans and MART, which is handled our proposed TinT Decoder. Notably, with the same action (i.e.,  run down the track and jump into a sand pit), our VLTinT can tell when the action starts (i.e., begin) and happens (i.e., then). This is thank to the rich spatial information of \encoder and strong temporal coherency of TinT Decoder. 

\begin{table*}[!hbt]
\centering
\caption{Ablation study on the contribution of each modality in \encoder on ActivityNet Captions dataset. Env., Agt., and Ling. denote the global visual environment, local visual main agents, and linguistic relevant scene elements, respectively.}

\resizebox{\linewidth}{!}{
\begin{tabular}{ccc|cccc|cc||cccc|cc}
\toprule
\label{tab:ablation_feat}
& & & \multicolumn{6}{c||}{\textit{at-test} split} & \multicolumn{6}{c}{\textit{ae-val} split}\\ \hline
Env. & Agt. & Ling. & B@4 $\uparrow$ & M $\uparrow$&  C $\uparrow$ & R $\uparrow$ &  Div@2$\uparrow$ & R@4 $\downarrow$ & B@4 $\uparrow$ & M $\uparrow$&  C $\uparrow$ & R $\uparrow$ & Div@2 $\uparrow$ & R@4 $\downarrow$ \\ \hline
\cmark & \xmark & \xmark & 13.62 & 17.41 & 29.09 & 35.96 & 76.14 & 5.97 & 14.02 & 17.58 & 30.31 & 36.20 & 76.11 &6.08  \\
\xmark & \cmark & \xmark & 11.83 & 16.22 & 21.39 & 33.97 & \underline{79.20} & \underline{4.16} & 12.13 & 16.57  & 24.98  & 34.36 & \underline{79.18}  & \underline{4.24}\\
\xmark & \xmark & \cmark & 13.38 & 17.69 & 30.30 & 35.63 & \textbf{80.50} & \textbf{3.32}  & 14.00 & 17.88  & 31.64  & 35.95 & \textbf{80.44}  & \textbf{3.22}\\ 
\cmark & \cmark & \xmark &  13.77 & 17.52 & 30.05 & 35.93 & 77.78 & 4.69 & 14.12 & 17.78  & 31.15  & 36.12 & 78.02  & 4.56\\ 
\cmark & \xmark & \cmark & \textbf{14.53} & \underline{17.79} & \underline{30.83} & \textbf{36.67} & 76.47 & 5.60 &
\underline{14.84} & \underline{17.97} & \underline{31.86} & \underline{36.80} & 76.41 & 5.67\\ 
\cmark & \cmark & \cmark & \underline{14.50} & \textbf{17.97} & \textbf{31.13} & \underline{36.56} & 77.72 & 4.75 & \textbf{14.93} & \textbf{18.16}  & \textbf{33.07}  & \textbf{36.86} & 77.72  & 4.87\\ \bottomrule
\end{tabular}
}
\end{table*}

\begin{table}[!t]
\centering
\caption{Performance comparison between two cases trained on TinT network without visual feature: (i) scene elements extracted by Mask R-CNN (M RCNN) (ii) scene elements extracted by CLIP.}
\begin{tabular}{l|cccc|c}
\toprule
 &  B@4$\uparrow$ & M $\uparrow$ &  C $\uparrow$ & R $\uparrow$ & R@4 $\downarrow$\\ \hline  
M RCNN & 1.35 & 9.09 & 12.29 & 23.06 & 18.52 \\ 
CLIP & \textbf{13.38 } & \textbf{17.69} & \textbf{30.30} & \textbf{35.63} & \textbf{3.32}\\ 
\bottomrule
\end{tabular}
\label{tab:concepts}
\end{table}

\begin{table}[!hbt]
\centering
\caption{Comparison between RNN and Transformer to model inter-event dependencies in TinT decoder on ActivityNet Captions with C3D (env+agent) is visual feature in the encoder. Linguistic feature (Ling.) is considered as an option.} 
\resizebox{\linewidth}{!}{
\begin{tabular}{ccc|cccc}
\toprule
\multirow{2}{*}{} & \multicolumn{1}{c}{use of} & \multicolumn{1}{c|}{inter-event} & \multirow{2}{*}{B@4 $\uparrow$} & \multirow{2}{*}{M $\uparrow$} & \multirow{2}{*}{C $\uparrow$} & \multirow{2}{*}{R $\uparrow$} \\ 
 & ling. &  modeling  &        &       &    & \\ \hline
\multirow{4}{*}{\rotatebox{90}{ae-val}}  
& \multirow{2}{*}{\xmark}  & RNN & 11.68 & 16.79 & 25.86 & 33.97  \\ 
& & Trans. & \textbf{14.12} &\textbf{ 17.78} & \textbf{31.15}  & \textbf{36.12}\\ \cline{2-7}
& \multirow{2}{*}{\cmark} & RNN  & 13.75 & 17.63  & 28.01  & 36.21  \\ 
& & Trans. &  \textbf{14.93}  & \textbf{18.16} & \textbf{33.07} & \textbf{36.86} \\ \midrule
\multirow{4}{*}{\rotatebox{90}{ae-test}}
& \multirow{2}{*}{\xmark}  & RNN  & 11.10 & 15.72 & 27.67 & 31.75 \\ 
& & Trans. &  \textbf{13.77}  & \textbf{17.52} & \textbf{30.05}  & \textbf{35.93}  \\\cline{2-7}
& \multirow{2}{*}{\cmark} & RNN  & 13.45 & 17.42 & 29.68 & 36.09 \\ 
& & Trans. & \textbf{14.50} & \textbf{17.97} & \textbf{31.13} & \textbf{36.56}\\
\bottomrule
\end{tabular}}
\label{tab:comp_rnn_trans}
\end{table}

\subsection{Quantitative Analysis}

We benchmark and compare \model with the prior SOTA VPC works on both ActivityNet Captions \textit{ae-val}, \textit{ae-test}, and YouCookII as in Tables. \ref{tab:anet_val}, \ref{tab:anet_test} and  \ref{tab:youcook}, respectively. In those tables, we highlight the \textbf{best} and the \underline{second-best} scores corresponding each metric. Compared to the SOTA approaches MART \cite{lei2020mart}, MART w/COOT \cite{ging2020coot}, and PDVC \cite{wang2021end}, our \model outperforms with large margins on both accuracy and diversity metrics on ActivityNet Captions. For example on \textit{ae-val} split, accuracy gains 3.13\%/1.56\%/5.80\%6.54\% on B@4/M/C/R metrics whereas diversity increases 0.32\% on Div@2 and reduces 0.32\% on R@4 compared to the second-best performance. On \textit{ae-test} split, accuracy gains 3.65\%/1.98\%/2.94\%5.71\% on B@4/M/C/R metrics whereas diversity increases 0.43\% on Div@2 and reduces 0.67\% on R@4 compared to the second-best performance. On YouCookII, our performance is the best on C, R, and R@4 metrics with considerable gaps while it achieves compatible performance on B@4 and M metrics. 

\subsection{Ablation Studies}

\vspace{2mm}
\noindent
\textbf{$\bullet$ Contribution of each modality in \encoder:} We examine VLTinT on ActivityNet Captions with different modality settings as given in Table \ref{tab:ablation_feat}. The first three rows show the performance on each individual modality whereas the last three rows show the performance on different combinations. Even though the best performance on overall is obtained by combining all three modalities of both vision (environment and agent) and language (scene elements), the performance on only linguistic feature is promising with notable performance, especially on diversity metrics. This should be included in our future investigation.







\vspace{2mm}
\noindent
\textbf{$\bullet$ Effectiveness of linguistic relevant scene elements:}
We compare the performance of VLTinT with two cases given in Table \ref{tab:concepts}: (i) scene elements extracted by Mask-RCNN trained on COCO with 80 classes \cite{MaskRCNN_ICCV17} and (ii) scene elements extracted by CLIP. The ablation study shows the effectiveness of the scene elements feature extracted by CLIP over Mask-RCNN. 
While scene elements consist of human/non-human (e.g., animals, vehicles) and visual/non-visual (e.g., relations, activities) elements, Mask R-CNN can only cover a small portion of them because it was trained on a small number of visual objects/classes, resulting in poor diversity and lower performance on scene understanding compared to CLIP. 


\vspace{2mm}
\noindent
\textbf{$\bullet$ Robustness of TinT Decoder: } We examine the TinT Decoder with two settings of inter-event modeling, i.e., RNN-based similar to \cite{lei2020mart} and  transformer-based (ours). 
The decoder is also considered with two encoder feature settings, i.e., with and without linguistic features whereas C3D (env+agent) is used as visual features. The result is shown in Table \ref{tab:comp_rnn_trans}. Here we observe the substantial performance gain by modeling inter-event relationships by our autoregressive outer transformer. 

\vspace{2mm}
\noindent
\textbf{$\bullet$ Effectiveness of VL Loss $\mathcal{L}_{VL}$: } The effectiveness of VL Loss is examined by replacing $\mathcal{L}_{VL}$ with MLE loss, which is a common loss in VPC. The performance of VLTinT on ActivityNet Captions \emph{ae-test} with two loss functions are reported in Table \ref{tab:l_clip}.

\vspace{2mm}
\noindent
\textbf{$\bullet$ Computational Complexity: } We compare computational complexity vs. accuracy of our VLTinT with SOTA VPC models on the ActivityNet \emph{ae-test} split. We report trainable params (millions), computation (GFLOPs), average inference time (seconds) over 100 random videos, and accuracy metrics in Table \ref{tab:complexity}. In this comparison, we investigate our VLTinT with different settings. Compared to SOTA, our model with only env. has compatible params and inference time with better performance, whereas our model with env. \& agent. \& lang. gain big margins on accuracy while the complexities remain plausible.




\begin{table}[t]
\centering
\caption{Effectiveness of $\mathcal{L}_{VL}$ compared to the standard MLE loss on ActivityNet Captions \emph{ae-test}. }
\begin{tabular}{l|cccc}
\toprule
Loss &  B@4$\uparrow$ & M $\uparrow$&  C $\uparrow$ & R $\uparrow$ \\\hline
MLE & 13.80 & 17.72 & 30.59 & 36.11  \\ 
$\mathcal{L}_{VL}$ (ours) & \textbf{14.50} & \textbf{17.97} & \textbf{31.13} & \textbf{36.56} \\ 
\bottomrule
\end{tabular}
\label{tab:l_clip}
\end{table}

\begin{table}[t]
\centering
\caption{Computational cost vs. accuracy between VLTinT with different settings and SOTA VPC models.}
\resizebox{\linewidth}{!}{\begin{tabular}{c|l|c|c|c||c|c}
\hline
 & & \multicolumn{3}{c||}{Computational cost} & \multicolumn{2}{c}{Accuracy} \\
 & Models & Params$\downarrow$ & Comp. $\downarrow$& Inf.$\downarrow$ & M$\uparrow$ & C$\uparrow$\\
 \hline
 & MART & 36.25 & 6.32 & 0.025 & 15.57 & 22.16 \\
 & Mem Trans & 29.69 & 256.44 & 0.706 & 16.10 & 27.36\\
 \hline
 \multirow{3}{*}{\rotatebox{90}{VLTinT}}& env & 36.01 & 17.69 & 0.028 & 17.41 & 29.09\\
 & env/agent & 40.37 & 22.70 & 0.032 & 17.52 &30.05 \\
 & env/agent/lang & 43.40 & 40.37 & 0.038 & 17.97 & 31.13\\
\bottomrule
\end{tabular}
}
\label{tab:complexity}
\end{table}

\section*{Conclusion}
In this work, we have presented \model, a novel model for VPC. The proposed network consists of \encoder and TinT Decoder. In \encoder, the video feature is extracted by three modalities, i.e., global visual environment, local visual main agents, and linguistic relevant scene elements; and they are fused through M2RF. In TinT Decoder, the intra-event coherency is modeled by the unified inner transformer and inter-event coherency is modeled by the autoregressive outer transformer. Our proposed \model is designed as an end-to-end framework and trained by our proposed VL contrastive loss $\mathcal{L}_{VL}$. 
Comprehensive experiments and extensive ablation studies on ActivityNet Captions and YouCookII datasets have demonstrated the effectiveness of \model, which outperforms the existing SOTA approaches on both accuracy (B@4, M, C, R) and diversity (Div@2, R@4) metrics. 

Future investigations might include further examining linguistic feature in video understanding and exploring the VL Encoder in other video analysis problems.

\textbf{Acknowledgments:}
This material is based upon work supported by the National Science Foundation (NSF) under Award No OIA-1946391, NSF 1920920, NSF FAIN-2223793 and NIH 1R01CA277739.

\footnotesize
\bibliography{main}

\end{document}